\documentclass[conference]{IEEEtran}
\IEEEoverridecommandlockouts
\usepackage{cite}
\usepackage{amsmath,amssymb,amsfonts}
\usepackage{graphicx}
\usepackage{textcomp}
\usepackage{xcolor}
\def\BibTeX{{\rm B\kern-.05em{\sc i\kern-.025em b}\kern-.08em
    T\kern-.1667em\lower.7ex\hbox{E}\kern-.125emX}}

\usepackage{adjustbox}
\usepackage{balance}
\usepackage{comment}
\usepackage{tabularx}
\usepackage{nth}
\usepackage{tikz}
\usepackage{varwidth}
\usepackage{multirow}

\usepackage{algorithm}
\usepackage{algpseudocode}

\usepackage{booktabs}

\usepackage{caption}
\usepackage{footnote}
\usepackage[flushleft]{threeparttable}

\usepackage{amssymb}

\usepackage{subcaption}
\usepackage{tcolorbox}

\usepackage{subcaption}
\makeatletter
\def\BState{\State\hskip-\ALG@thistlm}
\makeatother

\makeatletter
\renewcommand*\env@matrix[1][*\c@MaxMatrixCols c]{%
	\hskip -\arraycolsep
	\let\@ifnextchar\new@ifnextchar
	\array{#1}}
\makeatother

\begin{document}

\title{Detection of Deployment Operational Deviations for Safety and Security of AI-Enabled Human-Centric Cyber Physical Systems  }

\author{\IEEEauthorblockN{Bernard Ngabonziza, Ayan Banerjee, Sandeep K.S. Gupta}
	\IEEEauthorblockA{
		\textit{Arizona State University}\\
		\{bngabonz, abanerj3, sandeep.gupta\}@asu.edu}

}

\maketitle

\begin{abstract}
	In recent years, Human-centric cyber-physical systems have increasingly involved artificial intelligence to enable knowledge extraction from sensor-collected data. Examples include medical monitoring and control systems, as well as autonomous cars. Such systems are intended to operate according to the protocols and guidelines for regular system operations. However, in many scenarios, such as closed-loop blood glucose control for Type 1 diabetics, self-driving cars, and monitoring systems for stroke diagnosis. The operations of such AI-enabled human-centric applications can expose them to cases for which their operational mode may be uncertain, for instance, resulting from the interactions with a human with the system. Such cases, in which the system is in uncertain conditions, can violate the system's safety and security requirements.	
	
	This paper will discuss operational deviations that can lead these systems to operate in unknown conditions. We will then create a framework to evaluate different strategies for ensuring the safety and security of AI-enabled human-centric cyber-physical systems in operation deployment. Then, as an example, we show a personalized image-based novel technique for detecting the non-announcement of meals in closed-loop blood glucose control for Type 1 diabetics. 
	
\end{abstract}


\maketitle

\IEEEdisplaynontitleabstractindextext

%
\IEEEpeerreviewmaketitle

\section{Introduction}
Human centric monitoring and feedback systems are increasingly being used in practical settings reaching a significant user base. Examples include autonomous driver assist systems, wearable sensor based health monitoring systems, gesture based communication interfaces, and medical control systems such as closed loop blood glucose control systems for Type 1 Diabetic subjects. The primary characteristics of these applications are they are cyber-physical systems. This is because they involve closed-loop collaboration between the human user and the machine. In addition, most of these systems have some component that includes an artificial intelligence mechanism. 

Consider the example of a closed-loop glucose control system also known as Artificial Pancreas (AP). The AP system is a closed-loop system with a continuous glucose monitor (CGM) sensor sensing glucose levels from the tissue fluid and sending it to an infusion pump. This pump has a control software that uses an adaptive intelligent algorithm to first predict the blood glucose level \verb*|30 mins| in the future and compute the current infusion rate to keep the future blood glucose level within normal limits. The insulin is then infused by the pump at a steady rate.This system is a cyber-physical system and is enabled by an AI-software.

\subsection{Human Centric CPS System Model}

\begin{figure}[!h]
	\centering
	\tcbox[colback=white, colframe=black,boxrule=0.5mm ]{\includegraphics[width=6cm]{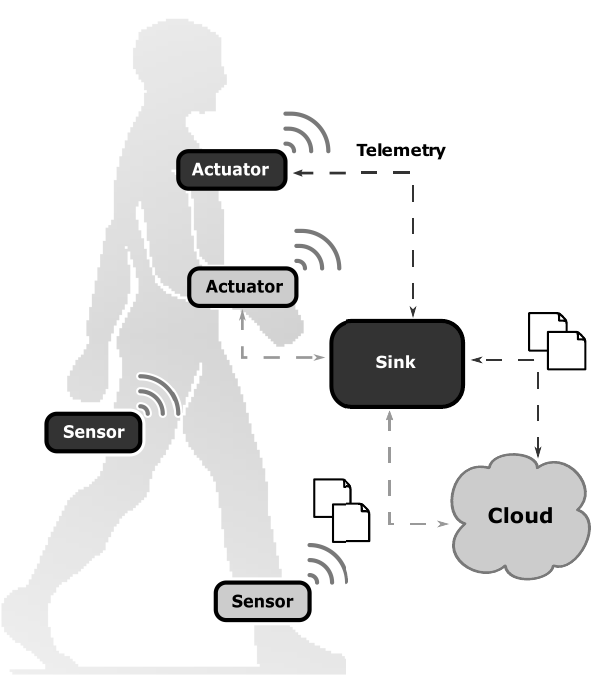}}
	\caption{System model}
	\label{fig:imageencode}
\end{figure}

Our system consists of several hardware devices (Fig. \ref{fig:imageencode}) and design layers . These layers consists of  
1) \textit{perception} (which gathers information and influence the action of the environment 
through sensing and actuation.), 2) \textit{network} (responsible for the communication between 
different devices), 3) \textit{service} (which provides various services, such as data abstraction
or running security protocols for the other three layers) and 4) \textit{application} 
( for interaction between the individual, stakeholder, and the system itself) layers. 

These systems contain a number of diverse, low-cost, wireless embedded 
{\em sensors} and a few {\em actuator} which together form a {\em distributed wireless network} 
around the individual  \cite{Kermani}. The sensors continuously monitor various physiological signals 
from the individual and wirelessly forward them to a {\em base station/sink} entity, usually implemented on a smartphone;  which is responsible for managing therapies, using the actuators present. The sink is also responsible for complex visualization, storage and forwarding the individual data to a cloud. 

\subsubsection{Sensors/Actuators}

The perception layer is responsible for influencing the environment and gathering information 
from through sensing and actuation. The main objective of the perception layer is to gain information from the environment and trigger some actions in response to the perceived information using sensors and actuators, respectively. These end devices are also called as \textit{nodes} in IoT-based systems.


\subsubsection{Sink}
Sensors can stream data to mobile phone or another control device via Blue-tooth in real-time. The mobile phones can host a set of control algorithms that determine the actuation inputs, or they can merely act as a data forwarder to the cloud. In some applications such as the Medtronic closed loop blood glucose control systems the controller and the actuator is combined into a single device. 

\subsubsection{Cloud Server}
The cloud server is data storage and computation hub. It is not only used as a computational and storage resource but also used as a knowledge resource in many AI applications. The large scale data repositories that are available in cloud hosted systems can be used to aide the development of predictive models.

\section{Problem Definition and Challenges of operational deviationDetection}

\subsection{Problem Definition}

An important characteristic of such systems is that since they are used in critical applications often these are not fully automatic. This means that certain components of the system require input from the human user. When the system is used in a way that it was not designed to be used, the system is entered into configurations that are untested or for which the verification result is uncertain can lead to potentially fatal safety violations. We would like to be able to detect this particular case.

To be very precise regarding the case of operational deviationwhich concerns us in this study. We will detect the change of operations which is defined as the case of Unknown|Unknowns by this article ~\cite{maity2023detection}. We will use a data driven approach in our study.

\subsection{Challenges and Trade-offs of operational deviation Detection}

The challenges that are associated with the detection of operational deviations is that it is very difficult to model this scenario and in addition it is a case in which the system has not been designed to operate. Moreover, The mixed approach of model based safety assurance and experimental analysis have several limitations. Experimental safety analysis is often expensive and hence can only be performed on a representative set of scenarios. It is an extremely difficult and often subjective task to select such a representative set that covers an exhaustive set of use cases that may occur in practice.

The solution is to supplement experimental analysis with model driven safety verification. It typically involves using a model to estimate the behavior of the system through the use of mathematical algorithms and simulation. The outcome is a set of parameter variations over time starting from an initial condition, a characteristic of the use case. This is often referred to as ``execution" of the system. These executions can then be compared with the safety condition to evaluate the safety of the system. The advantage of this approach is that simulations or mathematical estimations are less expensive and faster. In addition the system can be analyzed for a set of initial conditions (potentially containing infinite use cases) at one go instead of iterating through them one by one. 

This approach has been extensively used for many AI enabled human computer systems such as closed loop blood glucose control, and autonomous cars. However, the general problem of model based safety verification is intractable and cannot be solved accurately in limited time. Researchers have used several methods to approximate the system behavior over time and derive what is often referred to as \textit{reach set}. It is the approximate set of executions of a system for a given set of initial conditions representing use cases. Since the reach set is an approximation, this implies that the estimated behavior of the system for certain use cases are uncertain.Thus if such use cases actually occur in practice then the system behavior can possibly result in unsafe conditions. The usual practice is to design the system such that the probability of occurrence of an unsafe condition due to uncertainty in verification result is minimized. However, even if the probability is low, still there is a possibility of actual occurrence of an uncertain use case leading to safety violations.

When designing operation change detection mechanism there are several trade-offs that are taken into consideration. Next are the trade-offs we considered.

\subsubsection{Privacy vs. Personalization} Personalization in context-aware systems often demands access to detailed user data. While more data can lead to better personalization, it can also introduce serious privacy concerns. Users may be hesitant to share personal data, fearing misuse or data breaches ~\cite{ziegeldorf2014privacy}.

\subsubsection{Generalization vs. Personalization} General models can function across diverse scenarios but might not capture individual user patterns. Conversely, personalized models tailored to individual user behaviors can heighten detection accuracy but might demand frequent updating and may falter when faced with unexpected user behaviors.

\subsubsection{Data Privacy vs. Model Efficiency} Gathering vast amounts of data can bolster model performance, but it poses significant privacy concerns, particularly for human-centric applications. Employing privacy-preserving methods can diminish the model's efficacy. Like for example when we don't have access to certain data because of privacy concern of this data.

\subsubsection{Complexity vs. Transparency (Explainability)} Complex models may promise higher accuracy but often lack transparency, which is crucial in safety-critical applications. Simplified models offer greater transparency but may compromise detection capabilities. For example, when an autonomous car malfunctions and we cannot know why it malfunctioned or what caused an accident.

\subsubsection{Autonomy vs. Human Oversight} While fully autonomous systems ensure rapid reactions, human oversight remains essential in ambiguous scenarios to avoid undesirable outcomes. For example, when the patient must take urgent action to save his life.

\subsection{Proposed Solution}

\subsubsection{Personalized operational deviation Detection Model}
In this paper, we first focus on the unique challenges brought about by the change of operation practical deployments of AI enabled cyber human systems, to guarantee their safety. we introduces a framework for detecting operational deviations in such Human-centric cyber-physical-systems, ensuring their robustness in dynamic and unpredictable environments. We take the example of detecting missed meal announcements in the case of a closed-loop blood glucose control system or artificial pancreas; when the patient takes a meal but does not announce it to the controller that results in there being no insulin injected; then evaluate the impact of these challenges. We propose an example solution of a personalized image-based pattern recognition unique technique to detect the change of operation. 

\subsubsection{Personalized Image-based rescue meal Detection}
We will explore how to detect when a type 1 patient has diabetes has not communicated to the external device he must receive insulin. Our approach to detecting the missed meal announcement is to encode all of the interrelationships between the glucose and insulin signals in an image and to use image detection methods to distinguish between which images the interrelationships between glucose and insulin comes from the normal data. or those who come to the data where the meal has not been announced.

\section{Personalized Data-driven Framework for Detecting operational deviation}

\subsection{Overview of the Proposed Solution}
\begin{figure}[!h]
	\centering
	\tcbox[colback=white,  colframe=black]{\includegraphics[width=7cm]{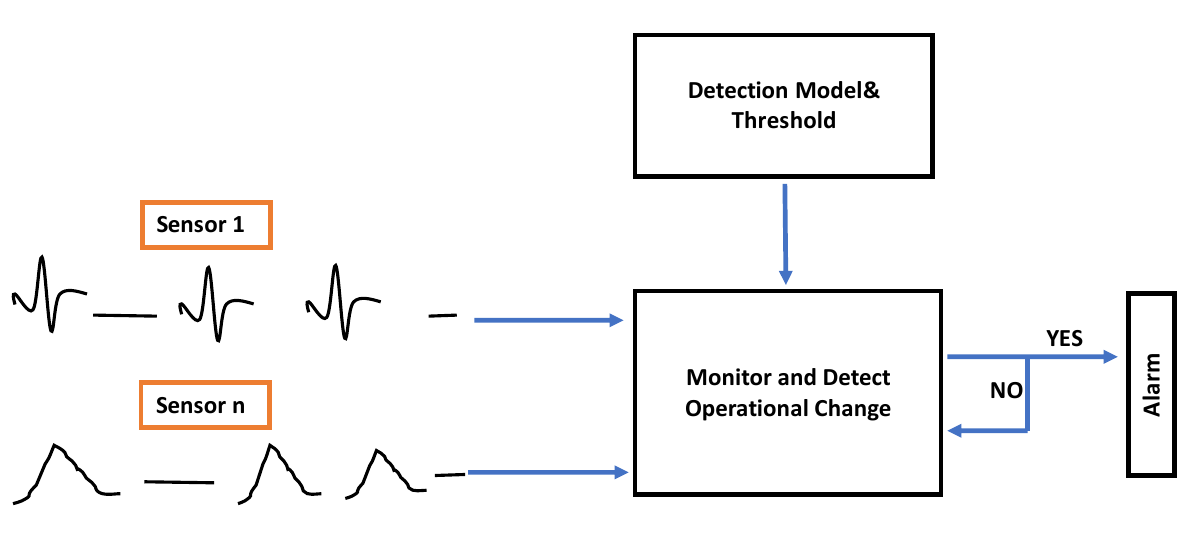}}
	\caption{Detection Architecture}
	\label{fig:Detectionarch}
\end{figure}

We will describe the components of the framework's architecture, how they work and interact with each other (Fig. \ref{fig:Detectionarch}). An operational deviation detection mechanism is a system that must detect the change in the functioning of the system and when this functioning deviates from the expected behavior, the mechanism must create an alert so that we can intervene to mitigate possible consequences. The proposed framework is made of several components. Here's how they works.

\subsection{Detection Model and Thresholds Rules} 

The first thing we do in designing the operational deviation detection mechanism is to create a detection model. We created a model using the historic data when the system is operating normally. Next, we must establish the threshold. The threshold determines that the system has deviated significantly from the rules that have been established. We establish the threshold and rule by statistical analysis of data and also the domain expertise of the system.

In our pursuit we will use data driven techniques to develop the model taking into account the threshold, to identify anomalies deviating from the behavior that we expect and which exceed the defined threshold.

\subsection{Monitoring and Detection of operational deviation} 

When the system comes into operation, critical variables are continuously monitored from data coming from sensors and other instruments that are part of the Cyber-physical-system. The data is collected and passed to the system monitor which will examine the data coming from the sensors and analyze it to detect any change in operation. When the change of operation has been detected, the alert is generated so that action can be taken to correct the error and prevent the harm from being caused to the person interacting with the system.

\section{Detecting Missed Meal-Announcement in Artificial Pancreas Systems}

The artificial pancreas is a system that must monitor the blood glucose level of the patient and administer insulin to control that blood glucose is maintained at the normal level. When a patient suffering from type 1 diabetes is fitted with an artificial pancreas, for example the Meditronic closed loop blood glucose control systems, they must inform the Meditronic system when he has swallowed something of nutritional value in calories and estimate how many calories there are. The problem arises when the patient does not informs the controller that he has eaten something resulting in the pump not injecting insulin into the patient. This same situation can be caused not necessarily by the fault of the patient but also by the malfunctioning of the controller or the pump itself.

This can compromise the effectiveness of the artificial pancreas, cause to be excess glucose in the patient's blood potentially causing hyperglycemia. It is for this reason that we must design methods to detect missed meal-announcement and ensure patient's safety. In our research we will find a way to detect when the patient with type 1 diabetes has eaten something but has not received insulin from the pump. 

\subsection{The Artificial Pancreas}

The artificial pancreas(AP) is used to help patients who suffer from type 1 diabetes. Type 1 diabetes is associated with the complete absence of insulin-secreting cells due to the immune-mediated destruction of such cells and results in severe hyperglycemia and in some patients ketoacidosis. These patients are treated with multiple daily insulin injections or an insulin pump. Such therapies may be used with the simultaneous use of a continuous glucose monitoring system (CGM). The CGM is an important component of daily diabetes management such use of complex insulin therapy without CGM is associated with hypo and hyperglycemia occurring daily. The use of a CGM and insulin pump provides the opportunity to automate insulin therapy with the ability to achieve glucose control that is closer to optimal. Such a system is referred to as a closed-loop glucose-insulin control system or artificial pancreas (AP).

The AP mainly consists of three systems, CGM for measuring glucose levels in the subcutaneous tissue, a control algorithm to calculate the amount of insulin that should be delivered, and a continuous SC insulin infusion (CSII) pump to deliver calculated insulin. The control algorithm conducts mathematical calculations to estimate the amount of insulin that has to be delivered but these algorithms are often augmented with manual patient-driven operations such as insulin boluses to handle meals or basal insulin. The artificial pancreas is constituted in our case by the insulin pump (which plays the role of the actuator) and the glucose sensor for the perception layer of our system model and in addition the controller. In this case, we will be using the Medtronic closed-loop blood glucose control systems, in which the controller and the actuator are combined into a single device.

\subsection{AI enabled CPS control systems :  AID System}

The AID system had these three components: a control algorithm running on a smartphone, this phone is connected wirelessly to the insulin pump and a Dexcom G6 CGM sensor.
The iAPS app that runs on unlocked Google Pixel 2 run the control strategy to compute the insulin
delivery as a function glucose velocity and insulin-on-board (IOB) ~\cite{gondhalekar2018velocity}.
The algorithm runs a zone-MPC algorithm which penalizes the deviation of the glucose level which is above the zone (90–120 mg/dL) during the day and (100–120 mg/dL) during the night. The computer insulin delivery rate is called micro bolus and is delivered every 5 minutes unless
any external event is detected. 

The external events can include: a) meal intake, where user utilizes a Bolus Wizard to compute a bolus input and manually command the system to
administer it. This must be in accordance with the participant’s carbohydrate ratio.
b) bolus correction, the system only allows a maximum of 2 U which can be added on the
meal bolus. This is when the glucose level is over 150 mg/dL ~\cite{dassau2017twelve}.

\subsection{Proposed Solution Workflow}

\begin{figure}[!h]
	\centering
	\tcbox[colback=white,  colframe=black]{\includegraphics[width=7cm]{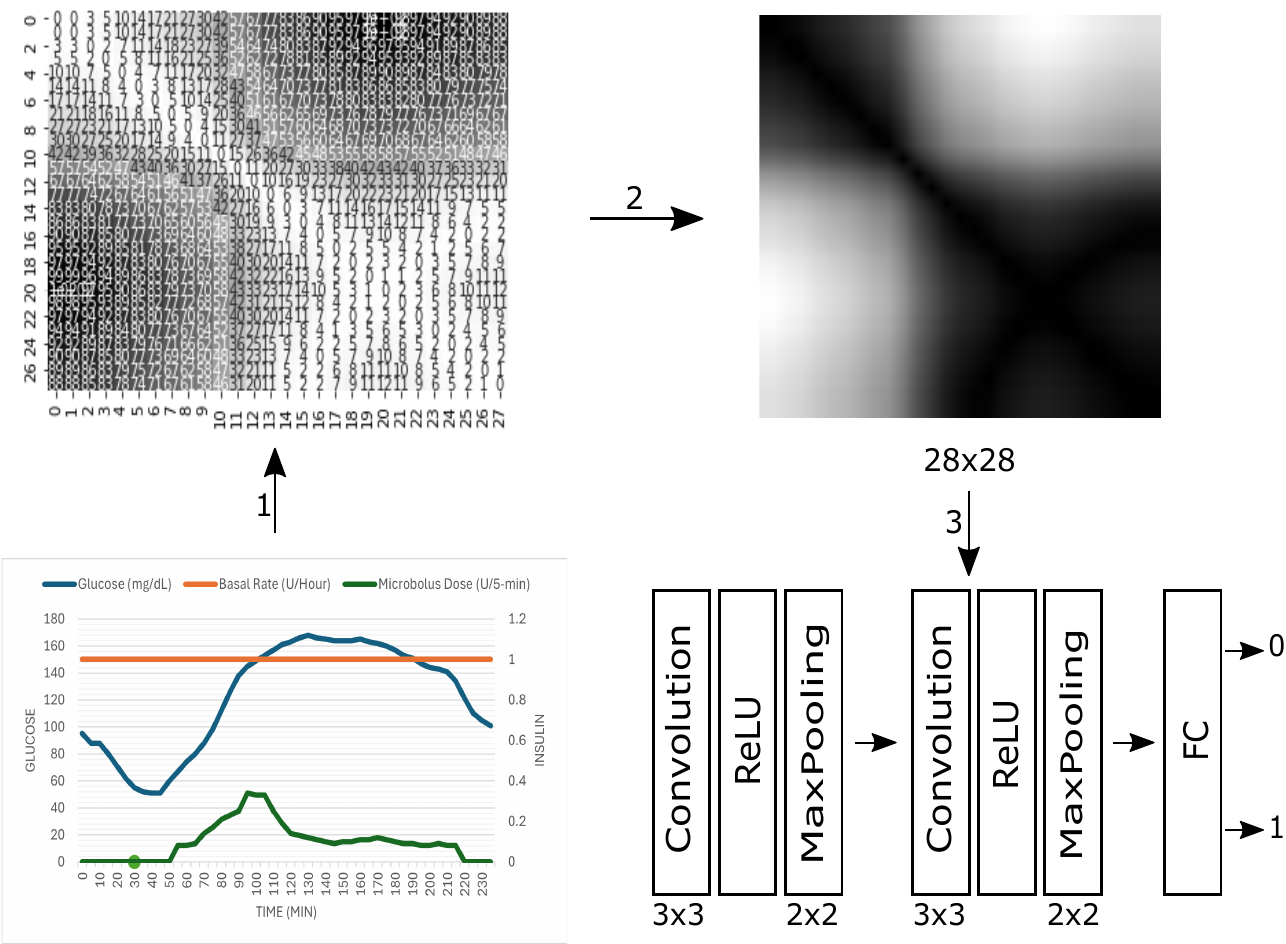}}
	
	\caption{Workflow}
	\label{fig:Workflow}
\end{figure}

Below is the workflow for our approach, it consists of three main steps which are marked by the arrows and the workflow (Fig. \ref{fig:Workflow}): First we have to convert the two time-series signals (CGM glucose and microbolus plus basal) into a distance matrix, then we transform this distance matrix into pixels of a grayscale image. The last step is to classify these images to determine if they came from a combination of insulin with the meal that was accompanied by the bolus or if it was the meal that was not announced and in this case, it was not there was no bolus taking.

\section{Experimental Results}

\subsection{DATASET}

The data that we used in this study were taken from the part of study which used the Automated insulin delivery (AID) system as described in this article ~\cite{kaur2022outpatient}. This study investigated the effects of psychological and pharmacological stress on glucose levels ~\cite{gonder2016psychological}. Participants in this study should be over 18 years old and had to have experience using an insulin pump for more than 3 months before being selected; also have an HbA1c <10.5\%.

In this study 14 patients who suffered type 1 diabetes were unrolled. However, only 12 were able to complete all stress induction sessions per protocol. 
During the study, breakfast constituted 20\%, lunch 30\%, and snack 15\% of all calories consumed per day. Of all the data collected over a period of 2 weeks. The snack data is the one of interest to all us, because this is the data that will help us calculate our problem because there is meal but no bolus. From the 12 patients, only 5 patients had enough snack event data (CGM, microbolus, and basal information) for the purpose of our study. We cut the CGM and insulin data where the patient took the snack. Precisely 30 minutes before the snack and 2 hours after the snack. The number of snack meals obtained from the five patients were 11, 11, 18, 7, and 8 respectively.

\begin{figure*}[!h]
	\centering
	\tcbox[colback=white,  colframe=black ,boxrule=0.5mm]{\includegraphics[width=13cm]{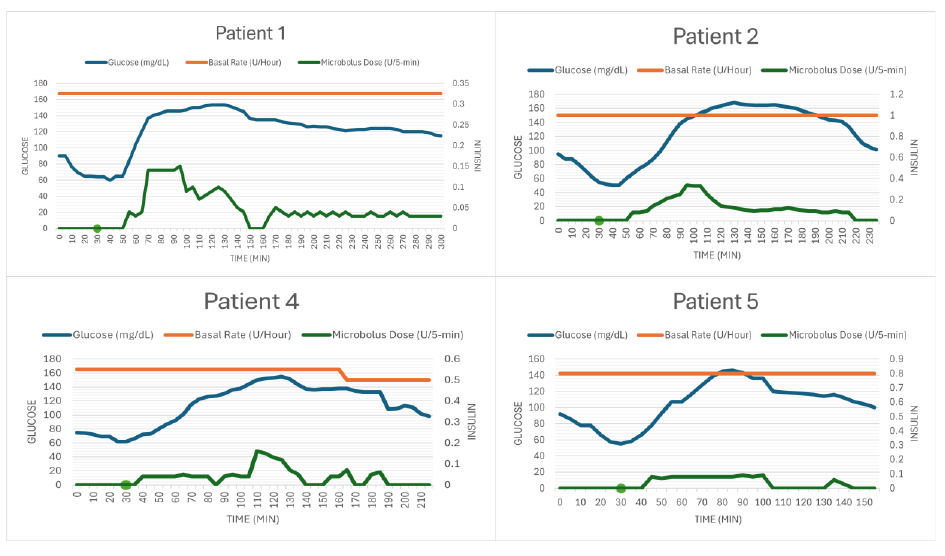}}
	\caption{Snack Data}
	\label{fig:dataplot}
\end{figure*}

These time series of data were used to encode the interrelations between glucose and insulin in an image using the algorithm described below. This portion of data will constitute the case that represents the rescue meal category for our machine learning classification model. Looking at the classification class that represents no meal, we consider the portion from 2 hours after the snack to 4 hours and 30 minutes after the snack (Fig. \ref{fig:dataplot}). After, use the same algorithm to convert these given time series into the images that will constitute the class category of our classifier for each patient.

In total, we collected 310, 290, 530, 210, and 215 images per class (rescue-meal, no-meal), for each patient respectively.

\subsection{Time-Series Image Encoding (Creating Image from time-series data)}



Glucose and insulin dynamics have been modeled since early 20th century, and thus many models have been proposed for the estimation of blood glucose. These mathematical models have been used to estimate the glucose disappearance and insulin sensitivity which are used to research on glucose-insulin dynamics and effects on blood glucose of insulin infusions. Michele et al. \cite{schiavon2014quantitative} ,adapting the miminal model \eqref{eq:1} by Caumo et. al \cite{caumo2000insulin}, proposed a insulin Sensitivity calculation  method from CGM and CSII. We adapted their calculations and created our Sensitivity-Relation matrix.

\subsection{Deriving the Interrelation (Sensitivity-Relation)}

The minimal model (Eqn. \ref{eq:1}) is described by the follow differential system of equation: 

\begin{equation}\label{eq:1}
	\begin{cases}
		\dot{g}(t) = -[p_1+x(t)]g(t)+p_1*g_b+\frac{r_a(t)}{v_g}  & \text{$g(0)=g_b$} \\ 
		
		\dot{x}(t) = -p_2*x(t)+p_3[i(t)-i_b]  & \text{$x(0)=0$}\\
		
	\end{cases}
\end{equation}\\

Where: \( G(t) \) (mg/dL) is the relative differential plasma glucose concentration, \( G_b \) (mg/dL) is the basal glucose concentration, \( X \) (unitless) represents the remote effects of insulin on glucose distribution and endogenous glucose production, \( I(t) \) (I/dL) is the blood insulin concentratio, \( I_b \) (I/dL) is the basal insulin concentration, \( P_1 \) (min\(^{-1}\)) is the glucose "mass action" rate constant, \( P_2 \) (min\(^{-1}\)) is the rate constant expressing the spontaneous decrease of tissue glucose uptake ability, \( P_3 \) (min\(^{-1}\)) is the insulin-dependent increase in tissue glucose uptake ability, per unit of insulin concentration excess over baseline insulin , \( V_G \) (dL/kg) is the insulin volume distribution, \( ra \) the glucose appearance following a meal.\\

Insulin Sensitivity is defined by the following equation which is the amount of glucose able to pass into the cells. 

\begin{equation}\label{eq:2}
	SI = \frac{p_3}{p_2}v_g 
\end{equation}

The insulin sensitivity \eqref{eq:2} was derived from this model \eqref{eq:1} by replacing  $p_1$ and  $x(t)$ by 
$\frac{GEZI}{v_g} + \frac{SI}{v_g}$ and $\frac{p_3}{p_2}\dot{x'}(t)$ respectively then solving for SI \\ 

Where: \( G_{EZ} \)  (min\(^{-1}\)) is the effect of glucose per se to increase glucose uptake into cells and lower endogenous glucose production at zero insulin.

We used their method to estimate over small time interval which involves solving the minimal model integral for which we want to estimate the insulin sensitivity and from the insulin sensitivity expression, we derived  our sensitivity relation. 

\begin{equation}\label{eq:3}
	SI_i= \frac{\int_{t_i}^{t_{i+d}}r_a(t)dt-GEZI\int_{t_i}^{t_{i+d}} \Delta g(t)dt -v_g \int_{t_i}^{t_{i+d}}\dot{g}(t)dt }{\int_{t_i}^{t_{i+d}}
		x'(t)g(t)dt+i_b\int_{t_i}^{t_{i+d}}\Delta g(t) dt}
\end{equation}\\

where d is the time interval we are calculating the sensitivity-relation for.

\subsubsection{Fraction of absorbed carbohydrate}

\begin{equation}\label{eq:4}
	\int_{t_i}^{t_{i+d}}r_a(t)dt = \frac{D.f(t_{i+d})-D.f(t_i)}{BW}
\end{equation}\\

Equation \eqref{eq:4} represents the amount of carbohydrate absorbed into plasma during the time interval 'd'
which we calculated according from Dalla's model \cite{dalla2006system}.
where $D.f(t)$ is the fraction ingested carbohydrate that has been absorbed into plasma.

\begin{equation*}\label{eq:5}
	SI_i= \frac{\frac{D.(f(t_{i+d})-f(t_i))}{BW}-GEZI.AUC( \Delta g)-v_g.(g(t_{i+d})-g(t_i))}{AUC(i)\frac{AUC(|\Delta g|)}{(t_{i+d}) - (t_i)}}
\end{equation*}\\

From equation \eqref{eq:5} we can solve for the area-under-the-curve relation $AUC_R$ between insulin and glucose.

\begin{equation}\label{eq:6}
	AUC_R= \frac{AUC(i)}{AUC(\Delta g)}= \frac{\frac{D.f(t_{i+d})-D.f(t_i)}{BW*AUC(\Delta g)}-GEZI - \frac{v_g (g(t_{i+d})-g(t_i))}{AUC( \Delta g)}}{ SI_i \frac{AUC(|\Delta g|)}{(t_{i+d}) - (t_i)}}
\end{equation}\\

The sensitivity-Relation matrix is defined as follow:
\begin{equation} \label{eq:12}
	SR= AUC_R( G(T),I(T)) 1<j,k<N 
\end{equation}

Given normalize CGM time series $g(t)$ and Basal Insulin signal $i(t)$, Sensitivity-Relation matrix $SR$, which is $NxN$ matrix.

Where SR, is the resulting Sensitivity-Relation matrix.

The algorithm for the Sensitivity-Relation matrix $AUC_R()$ is below (Alg. \ref{alg:SR}).


\begin{algorithm} 
	\caption{Sensitivity-Relation Creation Function }\label{euclid}
	\begin{algorithmic}[1]
		\State $ def \textit{ create\_matrix(x,y):} $ 
		\State \hskip1.5em $ x = np.asarray(x) $
		\State \hskip1.5em $ y = np.asarray(y) $
		
		\State  \hskip1.5em$\textit{m = len}\textit{(x) }$
		
		\State \hskip1.5em $result = np.empty((m,m),dtype=float)$  
		
		\State \hskip1.5em $for \textit{ i} \textit{ in } range(m):$
		\State \hskip1.5em \hskip1.5em  $ result[i,:] = AUC_R(x[i],y)$
		\State \hskip1.5em $ return \textit{ result} $

	\end{algorithmic}
	\label{alg:SR}
\end{algorithm}


\section{Time-Series Images Classification}

After generating images encoding calculated from Glucose and Insulin measurements for every patients, we used convolution neural network to classify which images belongs to which patient and which images correspond to tempered time series for example Glucose data and Insulin that do not belong to the same patient. \\

\subsection{CNN architecture }

Our architecture (Table \ref*{table:cnnarch})consists of two convolution layers (conv) , 2 max polling(maxp) layers and 2 fully connected layers(FC). Convolution layers extract specific features from the images and capture the relationship between the two signal, which has been encoded in the images using 
the recurrence plot.
Then to reduce the dimension of the feature extracted by the convolution layers, we pass the
their output to the polling layers and after the reduce feature which are 3D are flattened to be be passed to the fully connected layer which will finally perform the classification using the softmax as the activation function.

\begin{table}[ht]
	\centering
	\caption{CNN Architecture}
	\label{table:cnnarch}
	\begin{tabular}{|c|c|c|c|c|}
		\hline
		Layer            & Filter & Kernel & Stride & Padding \\ \hline
		Conv2D+ReLU      & 32              & 3           & 1      & 0      \\ \hline
		MaxPool2D     	 & -              & 2           & 1      & 0       \\ \hline
		Conv2D+ReLU      & 16              & 3           & 1      & 0       \\ \hline
		MaxPool2D     	 & -              & 2           & 1      & 0       \\ \hline
		Flat+FC1+ReLU    & 512             & -           & -      & -       \\ \hline
		FC2              & 2               & -           & -      & -       \\ \hline
	\end{tabular}
\end{table}

\section{Training and Inference}

For training and testing,  we used a 30 percent for training / testing split. We performed prediction on the test set and report  performance in terms of the following metrics. 

Let TP denote the number of True Positives, FP as False Positives, TN as True Negatives and FN as False Negatives obtained for each instance among the 2 labels considered.

\begin{equation}	Accuracy = 	 \frac{TP+TN}{TP+TN+FP+FN}  	\label{eqn:accur}	\end{equation} 
\begin{equation}	Precision =  	 \frac{TP}{TP+FP} 	\end{equation}
\begin{equation}	Recall = 	 \frac{TP}{TP+FN}	\end{equation}
\begin{equation}	F1 =	  \frac{2*Precision*Recall}{Precision+Recall} = \frac{2*TP}{2*TP+FP+FN}  \label{eqn:f1}	\end{equation} \\

The  summary of results for all the patient are in the table and figures below.

\begin{figure}[!h]
	\centering
	{\includegraphics[width=9cm]{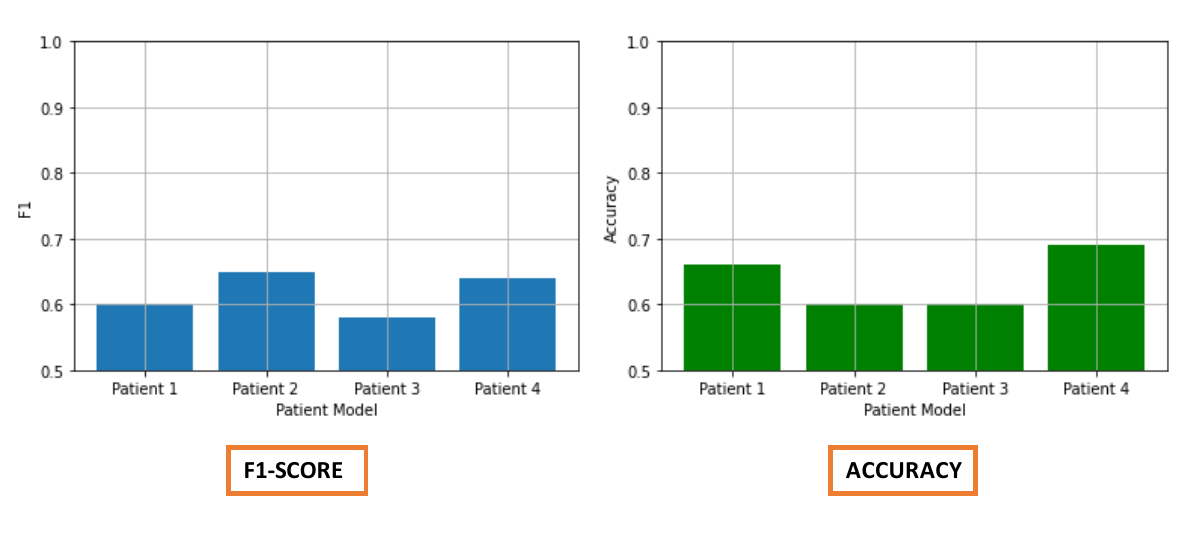}}
	\caption{Accuracy and F1-Score}
	\label{fig:dataexlpoler}
\end{figure}

In the table above is the result of experience regarding detecting missed meal-announcement for our five patients. The table contains the accuracy (Eqn \ref{eqn:accur}) and F1 score (Eqn \ref{eqn:f1}) for five different personalized model for each patient, labeled as "Patient 1", to "Patient 5" respectivel (Table \ref*{table:accf1}) . The value for accuracy and F1 scores has been rounded to two decimal places ( Fig \ref{fig:dataexlpoler}).

\begin{table}[!h]
	\begin{tabular}{|c|c|c|c|}
		\hline
		Patient & Our Model Accuracy & Our Model F1 & VIT Model ~\cite{zhu2023understanding} \\ \hline
		
		1 & 0.66 & 0.60 &0.63 \\ \hline
		2 & 0.60 & 0.65& 0.60 \\ \hline
		3 & 0.60 & 0.58  &0.54  \\ \hline
		4 & 0.69 & 0.62 &0.45 \\ \hline
		5 & 0.64 & 0.62&0.50 \\ \hline
		
	\end{tabular}
	\centering
	\caption{Accuracy and F1 Scores}
	\label{table:accf1}
\end{table}

\section{Conclusion}

This study showed us that the accuracy and the F1 score vary patient to patient, As intra-patient variability influences the accuracy of the detection system. Hence our use of personalized model to solve the problem of detect missed meal-announcement. Even if the model for detect the missed meal announcement achieved good accuracy and F1 scores for some patients, for others it performed less well. Nevertheless, this study gave us promising results in the method of converting the interrelation of glucose and insulin into image and then using machine learning techniques on images which are powerful tools for data processing. Which also shows us the potential of our method to detect operational deviation in deployed multi-input AI-enabled systems. In the future we will explore how to make our detection model more explainable and transparent, to make sure that the model is able to adapt to new input, and that stakeholders understand the rational behind the detection method of the change in operation. 
We are going to adapt our model in the scenario of autonomous cars, we will first obtain the data, then compose what constitutes the change in operation. Finally, we will extract the interactions between the signals of interest and encodes these signals into images, to detect the changes in operation. 

\bibliographystyle{IEEEtran}
\bibliography{Refsecon}
\end{document}